\documentclass[letterpaper, 10 pt, conference]{ieeeconf}  
\let\labelindent\relax

\IEEEoverridecommandlockouts                              

\title{\LARGE \bf
OpenGS-SLAM: Open-Set Dense Semantic SLAM with 3D Gaussian Splatting for Object-Level Scene Understanding
}

\author{Dianyi Yang$^{1,2}$, Yu Gao$^{1,2}$, Xihan Wang$^{1,2}$, Yufeng Yue$^{1,2}$, Yi Yang$^{*, 1,2}$, Mengyin Fu$^{1,2}$ 
\thanks{*This work was partly supported by National Key R\&D Program of China (2022YFC2603600) and National Natural Science Foundation of China (Grant No. NSFC 62233002)}
\thanks{$^{1}$School of Automation, Beijing Institute of Technology, Beijing, China}%
\thanks{$^{2}$National Key Lab of Autonomous Intelligent Unmanned Systems,
Beijing Institute of Technology, Beijing, China}%
\thanks{*Corresponding author: Y. Yang Email: yang\_yi@bit.edu.cn}%
}

\usepackage{graphicx} 
\usepackage{float} 
\usepackage{subfigure} 
\usepackage{cite}
\usepackage{color}
\usepackage{url}
\usepackage{booktabs}
\usepackage{hyperref}
\usepackage{footnote}
\usepackage{makecell}
\usepackage{amssymb}
\usepackage{hyperref}
\usepackage{threeparttable}
\usepackage{colortbl}
\usepackage{xcolor}
\usepackage{tabularx}
\usepackage{pgfplotstable} 
\usepackage{colortbl}      
\usepackage{multirow}
\usepackage{enumitem}

\usepackage{threeparttable}
\usepackage{amsmath}
\usepackage{caption}

\definecolor{lightgreen}{rgb}{0.7176, 0.8902, 0.7216}
\definecolor{lightyellow}{rgb}{0.9843, 0.9765, 0.7020}

\begin{document}

\newcommand{\colorcell}[1]{%
  \pgfmathparse{min(100,max(0,(#1/0.5)*100))}  
  \xdef\tempa{\pgfmathresult}  
  \pgfmathtruncatemacro\redcomponent{\tempa}  
  \pgfmathtruncatemacro\bluecomponent{100-\tempa}  
  \edef\tempb{\noexpand\cellcolor{red!\redcomponent!blue!\bluecomponent}}  
  \tempb #1  
}

\maketitle
\thispagestyle{empty}
\pagestyle{empty}

\begin{abstract} Recent advancements in 3D Gaussian Splatting have significantly improved the efficiency and quality of dense semantic SLAM. However, previous methods are generally constrained by limited-category pre-trained classifiers and implicit semantic representation, which hinder their performance in open-set scenarios and restrict 3D object-level scene understanding. To address these issues, we propose OpenGS-SLAM, an innovative framework that utilizes 3D Gaussian representation to perform dense semantic SLAM in open-set environments. Our system integrates explicit semantic labels derived from 2D foundational models into the 3D Gaussian framework, facilitating robust 3D object-level scene understanding. We introduce Gaussian Voting Splatting to enable fast 2D label map rendering and scene updating. Additionally, we propose a Confidence-based 2D Label Consensus method to ensure consistent labeling across multiple views. Furthermore, we employ a Segmentation Counter Pruning strategy to improve the accuracy of semantic scene representation. Extensive experiments on both synthetic and real-world datasets demonstrate the effectiveness of our method in scene understanding, tracking, and mapping, achieving 10× faster semantic rendering and 2× lower storage costs compared to existing methods. Project page: \href{https://young-bit.github.io/opengs-github.github.io/}{https://young-bit.github.io/opengs-github.github.io/}.

\end{abstract}

\section{INTRODUCTION}

Dense semantic Simultaneous Localization and Mapping (SLAM) is a fundamental challenge in robotics\cite{robotics1,robotics2} and embodied intelligence\cite{embodied1, embodied2}. It integrates semantic understanding of the environment into 3D scene reconstruction and estimates camera pose simultaneously. Traditional semantic SLAM struggles with predicting unknown areas and requires significant map storage \cite{traditional}. While NeRF-based methods \cite{fusion,sni,NIDS,DNS} mitigated these issues, they still suffer from inefficient per-pixel raycasting rendering \cite{splatam}.

Alternative to NeRFs, the recently emerged 3D Gaussian Splatting (3DGS)\cite{3dgs} has shown remarkable reconstruction quality with high training and rendering efficiency. Following these advantages, 3DGS-based Semantic SLAM methods\cite{SemGaussSLAM,NEDS} are developed to achieve high-quality semantic mapping. These methods embed 2D semantic features into 3DGS representation, where each Gaussian is enhanced with an N-channel feature embedding.

However, these methods often rely on pre-trained classifiers with a limited number of categories to produce 2D segmentation results, which limits their effectiveness in open-set scenarios. Further, since all semantic information is implicitly stored within the feature-embedded representation, it becomes challenging to directly access the semantic label of each Gaussian. This makes them incompatible with embodied intelligence that necessitates 3D object-level scene understanding and interaction.

\begin{figure}
    
    \centering
    \includegraphics[width=0.48\textwidth]{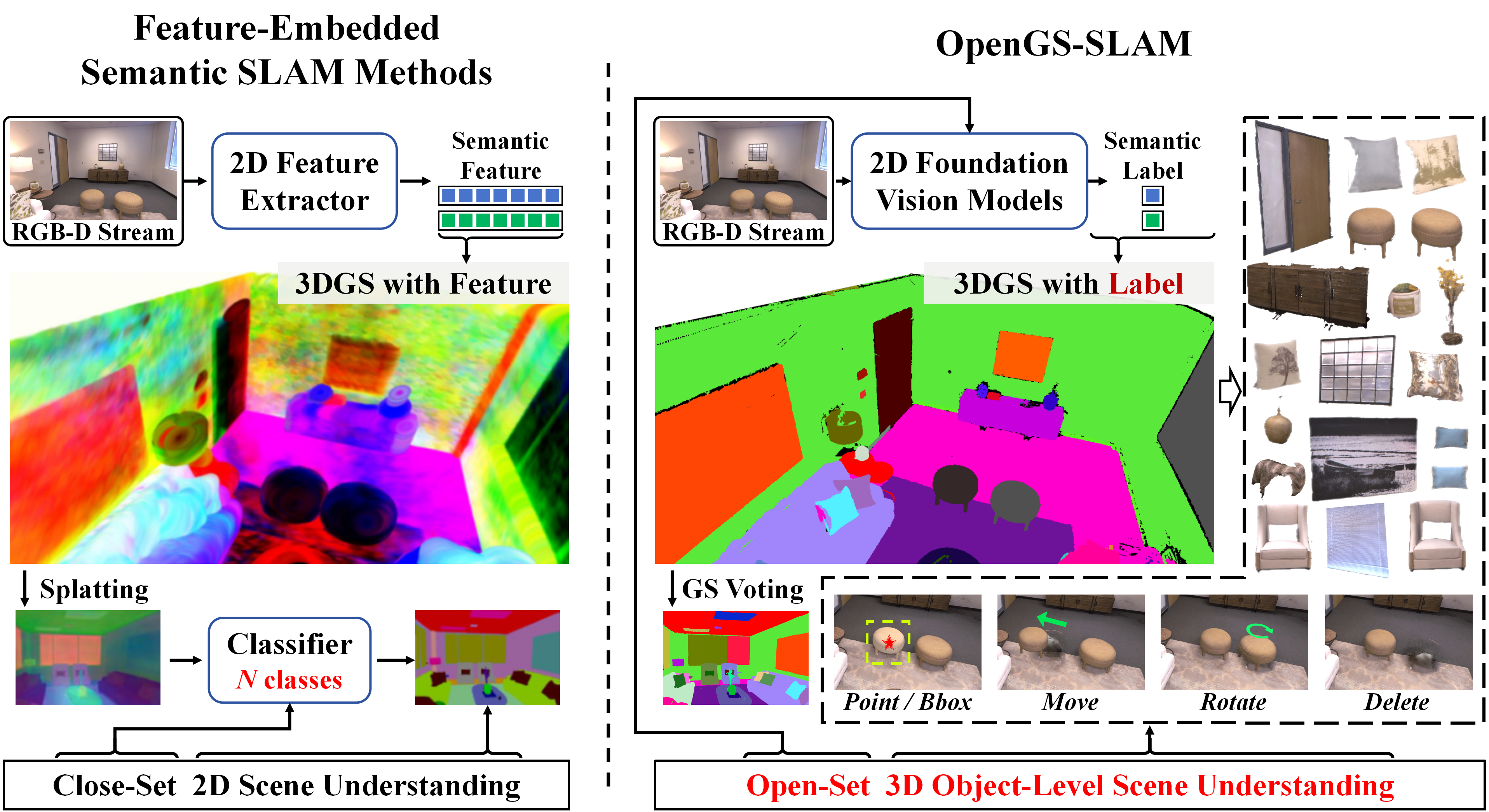}
    \caption{Compared to the feature-embedded methods\cite{SemGaussSLAM,NEDS}, our approach integrates semantic labels into the 3D Gaussian scene representation, ensuring that Gaussians belonging to the same object are consistently labeled. This enables more effective 3D object-level scene understanding and interaction. By leveraging 2D foundational vision models, our approach facilitates open-set dense semantic SLAM. The images on the left are from \cite{SemGaussSLAM}.}
    \label{fig1}
    \vspace{-1.0em}
\end{figure}

To address these limitations and extend semantic SLAM to open-set scenarios, we propose OpenGS-SLAM, a novel framework that incorporates 2D foundational vision models and assigns an explicit semantic label to each Gaussian, while enabling online semantic mapping. The foundational model in this framework is easily customizable; for example, it can be replaced with SAM \cite{SAM1} or MobileSAMv2\cite{Mobilesamv2} to ensure flexibility. Additionally, the framework can map semantic labels generated by these models into 3D Gaussians, facilitating its application in open-set scenarios. By assigning an explicit semantic label to each Gaussian, the framework efficiently supports 3D object-level scene understanding.

During the development of this framework, we faced three major challenges: \textbf{1)}  the non-differentiable nature of the semantic label attribute prevents us from using 3DGS rasterization for fast label map rendering and scene updating; \textbf{2)} inconsistencies in SAM’s results across multiple views cause the same object to receive varying labels or be segmented into parts or wholes; and \textbf{3)} during training, Gaussians in regions with less view constraint may expand significantly, resulting in inaccurate segmentation.

Focusing on these challenges, \textbf{1)} we introduce Gaussian Voting Splatting (GS Voting) to facilitate rapid 2D label map rendering and scene updating. \textbf{2)} To achieve consistent segmentation and efficiently incorporate new semantic information into the scene, we propose a Confidence-based 2D Label Consensus method. \textbf{3)} We propose a Segmentation Counter Pruning strategy to achieve more accurate segmentation of the scene. Additionally, we designed an ensemble semantic information generator to produce 2D semantic labels.

To summarize, our contributions are as follows:
   
\begin{itemize}[leftmargin=*]
    \item We propose OpenGS-SLAM, which enables open-set semantic SLAM based on 3D Gaussian Splatting. It supports a wide range of off-the-shelf 2D foundational vision models without extra training.

    \item We propose a novel semantic 3DGS representation with the Gaussian Voting Splatting method, which enables 3D object-level scene understanding with \textbf{10× faster} rendering from novel views, \textbf{2× lower}  storage costs and more accurate semantic segmentation compared to other methods.

    \item Our method achieves competitive performance in both synthetic and real-world scenarios in terms of tracking, mapping, and scene understanding.
\end{itemize}

\section{Related Work}

\textbf{Gaussian Splatting based RGB-D SLAM.} 3D Gaussian Splatting\cite{3dgs} revolutionized SLAM by enabling high-fidelity scene reconstruction and precise pose estimation. Notable advances include SplaTAM\cite{splatam} with silhouette-guided optimization, GS-SLAM's\cite{gs-icp} adaptive Gaussian expansion, and GS-ICP SLAM’s G-ICP for accurate pose tracking. Additionally, 3DGS has been extended for semantic scene understanding, with methods like SGS-SLAM\cite{SGS} incorporating 2D semantic priors. SemGauss-SLAM\cite{SemGaussSLAM} and NEDS-SLAM\cite{NEDS} further optimize memory usage through embedding low-dimensional semantic vectors into the 3D Gaussian framework.  However, the reliance on auxiliary classifiers or lightweight encoders for 2D segmentation limits their performance in open-set scenarios.

\textbf{2D Foundation Vision Models.} These methods are designed to capture broad knowledge from large-scale data and perform zero-shot tasks on unseen data. The Recognize Anything Model\cite{ram} can generate open-set classes from images. Segment Anything\cite{SAM1} excels in zero-shot segmentation, with advancements like MobileSAMv2\cite{Mobilesamv2} achieving over 20 fps in full-scene segmentation. SAM2.0\cite{sam2} extends to real-time video segmentation but struggles with new object detection and multi-object tracking. Methods like Yolo-world \cite{Yolo} and Grounding Dino\cite{grouding} enable zero-shot object detection. Some approaches\cite{Grounded-SAM, semantic-sam} integrate these models for zero-shot semantic segmentation.

\textbf{Open-World 3DGS Scene Understanding Method.} Recent methods in this field focus on lifting information from 2D foundation models to build 3D representations in an offline manner. For instance, LangSplat\cite{Langsplat}, Language-Embedded\cite{Language_embedded}, and Feature-3DGS\cite{Feature-3dgs} leverage CLIP\cite{clip} and SAM\cite{SAM1} for 2D pixel-level understanding. In contrast, GS-Grouping\cite{Gaussian-Grouping} and OpenGaussian\cite{OpenGaussian} elevate SAM’s 2D segmentation for 3D segmentation, with GS-Grouping using DEVA\cite{DEVA} to resolve frame-to-frame inconsistencies. However, these methods rely on time-consuming processes, limiting their applicability for online reconstruction.

\begin{figure*}
    \vspace{0.5em}
    \centering
    \includegraphics[width=0.99\textwidth]{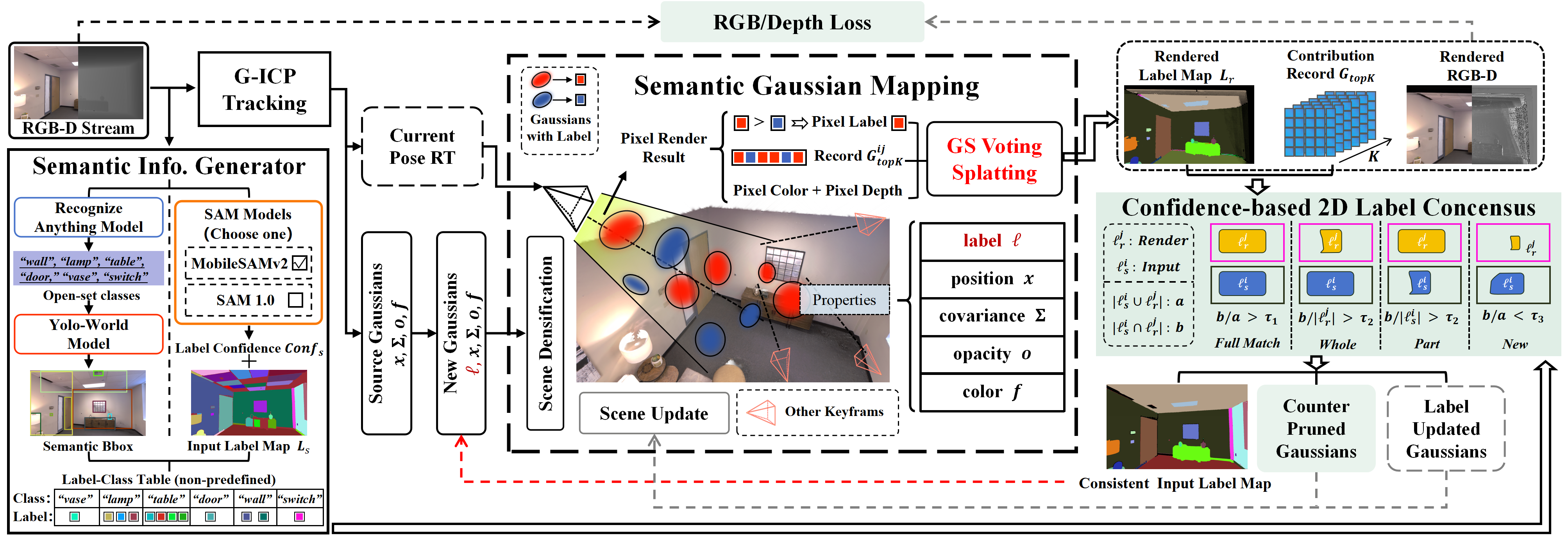}

    \caption{\textbf{An overview of OpenGS-SLAM.} Our method takes an RGB-D stream as input. RGB images are first processed by the Semantic Information Generator and G-ICP to extract semantic information and estimate the current pose. Using this pose, we perform precise and efficient semantic rendering via Gaussian Voting Splatting. We then unify the input label map with the current map through Confidence-based 2D Label Consensus, ensuring semantic consistency. During this process, partial Gaussian data is updated, and counter Gaussians are pruned.}
    
    \label{fig2}
    \vspace{-1.0em}
\end{figure*}

\section{Method}

\subsection{Ensemble Semantic Information Generator}

To fully exploit the capabilities of existing 2D foundational models in open-set scenarios, we integrate three types of expert models designed for specific contexts. The input RGB images are first fed into a Recognize Anything Model (RAM)\cite{ram} and a SAM model. The RAM generates open-set classes information, which serves as a text prompt for the Detection Model (using YOLO-World\cite{Yolo}) to produce semantic bounding boxes and detection scores. The SAM model outputs a label map with confidence scores, though the labels are randomly assigned. For each label, we match the bounding box with the highest IoU (above 0.5) to establish semantic correspondence, which is recorded in an input Label-Class Table \( \mathcal{M}_{LC}^{I} \). We also maintain a global Label-Class Table \( \mathcal{M}_{LC}^{G} \) throughout the SLAM process. For labels without a matching semantic bounding box, we assign them to the \textit{None} class with a detection score of 0.0. The expert models are customizable; for instance, we test SAM1.0\cite{SAM1} and MobileSAMv2\cite{Mobilesamv2} in the experimental section to assess their performance and adaptability in various scenarios. Notably, the categories in these tables are not predefined; they begin from an empty set and can expand dynamically during the SLAM process.

\subsection{Semantic Gaussian Mapping}
In this section, we first introduce our scene representation strategy. Next, we explain how to render both RGB-D images and label maps using Gaussian Voting Splatting. Finally, we describe the process of scene densification and updating.

\textbf{3D Gaussian strategy.} 
We employ a set of Gaussians with novel designed properties for scene representation. Compared with all previous methods, we replaced the feature embeddings in each Gaussian with a one-dimensional, non-differentiable attribute named \textit{GS Label}. Specifically, each Gaussian includes a 3D center position $x$, covariance \(\Sigma\), opacity $o$, color $f$, and label $\ell$.

\textbf{Gaussian Voting Splatting.}  
This method combines the strengths of 3D Gaussian Splatting and our semantic 3DGS scene representation to achieve high-fidelity RGB-D rendering alongside precise semantic mapping.
First, with the optimized 3D Gaussian scene representation, each Gaussian is projected from 3D space onto the 2D image plane using the camera pose \( T = \{R, t\} \). The covariance matrix \(\Sigma\) is transformed to the 2D plane as:
\begin{equation}
\Sigma' = J T^{-1} \Sigma T^{-T} J^T,
\end{equation}
where \(J\) represents the Jacobian of the projection function. Second, the Gaussians are sorted by depth and rendered using front-to-back alpha compositing. The pixel color \(C(p)\) and depth \(D(p)\) are calculated through \(\alpha\)-blending, as denoted in Eq.\ref{blending},
where \(f_i\) and \(d_i\) denote the color and depth of the \(i\)-th Gaussian, respectively, and \(\alpha_i\) is the opacity-weighted density, derived from the 2D covariance \(\Sigma'\) and opacity \(o_i\).

\begin{equation}
\left\{
\begin{aligned}
C(p) = \sum_{i \in N} f_i \alpha_i \prod_{j=1}^{i-1} (1 - \alpha_j), \\
D(p) = \sum_{i \in N} d_i \alpha_i \prod_{j=1}^{i-1} (1 - \alpha_j)
\end{aligned}
\label{blending}
\right.
\end{equation}

Third, for each pixel \(p\) in the final rendered label map, we aggregate all contributing Gaussians, denoted as \(G_p\), and apply label-aware \(\alpha\)-blending to determine the contribution of each label. Specifically, let \(w_j^i\) represent the weight of the \(i\)-th Gaussian \(g_i\) with label \(\ell_j\) in \(G_p\), computed as:
\begin{equation}
    w_j^i = \alpha_i \prod_{k=1}^{i-1} (1 - \alpha_k)
\end{equation}
where \(k\) refers to the depth-sorted index of Gaussians in \(G_p\). Defining \(G_p^j\) as the set of Gaussians in \(G_p\) with \(\ell_j\), the total weight \(W_j\) for \(\ell_j\) is computed as:
\begin{equation}
    W_j = \sum_{g_i \in G_p^j} w_j^i.
\end{equation}

The final label assigned to pixel \(p\) corresponds to the label with the highest cumulative weight.
This approach ensures that each pixel is labeled based on the most significant cumulative contribution from the Gaussians, providing accurate and efficient 2D label map rendering.

\textbf{Scene Densification and Updating.} For densification, once a frame is identified as a keyframe, we use the covariance computed from G-ICP to initialize the source Gaussians\cite{gs-icp}. After integrating the semantic \textit{GS Label} attribute, these Gaussians are added into the map.

For updating, we optimize the differentiable parameters using a loss function applied to selected frames following \cite{gs-icp}.
In contrast, the non-differentiable \textit{GS Label} is updated in an explicit manner. During the label map rendering, we record the top \( K = 50 \) Gaussians contributing to each pixel, generating the matrix \( G_{topK} \in \mathbb{R}^{H \times W \times K} \). If a pixel \( p \) labeled as \( \ell_s \) in the rendered label map, is updated to \( \ell_t \), we utilize \(  G_{topK} \) to identify the Gaussians previously labeled as \( \ell_s \) that contribute to pixel \( p \) and update their label to \( \ell_t \). This approach ensures efficient and precise updates of semantic labels across the scene.

\subsection{Confidence-based 2D Label Concensus}
\label{sec:concensus}
 Currently, models like MobileSAMv2\cite{Mobilesamv2} that support real-time full-scene segmentation struggle with multi-view inconsistencies, where the same object can end up with different labels or be segmented into parts/whole across different viewpoints. 
 Most existing methods address this through additional tracking models\cite{Gaussian-Grouping} or global maps\cite{Sai3d,Feature-3dgs}, limiting their real-time applicability. Therefore, we propose a Confidence-based 2D Label Consensus method, leveraging the efficient 3DGS scene rendering to unify input semantic information with the current map on-the-fly.
 
 We define \( \mathcal{L}_s = \{\ell_s^i\}_{i=1}^M \) as the set of labels in the input label map, with \( \mathcal{C}_s \) representing the confidence of each label. Similarly, \( \mathcal{L}_r = \{\ell_r^j\}_{j=1}^N \) denotes the set of labels in the rendered label map from the current viewpoint, with \( \mathcal{C}_r \) as the corresponding confidence. Additionally, we define \( |\ell| \) as the number of pixels associated with label \( \ell \) in its respective label map. Our goal is to achieve label consensus by matching all labels in \( \mathcal{L}_s \) with those in \( \mathcal{L}_r \), or creating new labels if necessary. The final output is a Consistent Input Label Map that provides consistent semantic information.
 
 For any \( \ell_s^i \in \mathcal{L}_s \) and \( \ell_r^j \in \mathcal{L}_r \), we identify four possible relationships based on their \( |\cdot| \) in the label map, as shown in Fig.\ref{fig2}, where we set three threshold as \( \tau_1, \tau_2, \tau_3 \).

 
\begin{itemize}[leftmargin=*]
    \item \textit{Full Match:} If \( \ell_s^i \) fully matches \( \ell_r^j \), we update the label in \( \mathcal{L}_s \) from \( \ell_s^i \) to \( \ell_r^j \) and set \(  \mathcal{C}_r^j\) to the maximum of its current value and \( \mathcal{C}_s^i \).
    
    \item \textit{Partial Match:} We define \( \mathcal{P}_s^j \in \mathcal{L}_s \) as all part labels that partially matches to \( \ell_r^j \). We then calculate their area-weighted average confidence \( \bar{\mathcal{C}}_s \), computed as:
    \begin{equation}
        \bar{\mathcal{C}}_s = \frac{\sum_{\ell_s^i \in \mathcal{P}_s^j} \mathcal{C}_s^i \times |\ell_s^i|}{\sum_{\ell_s^i \in \mathcal{P}_s^j} |\ell_s^i|}
    \end{equation}
    If \( \mathcal{C}_r^j > \bar{\mathcal{C}}_s \), we update all part labels in \(\mathcal{P}_s^j \) to \( \ell_r^j \). Otherwise, new label categories are assigned to those \(\ell_s^i\)  whose \( \mathcal{C}_s^i > \mathcal{C}_r^j \). For each new label assignment \( \ell_s^i \to \ell_{\text{t}} \), we search the corresponding region in \(  G_{topK} \) for all Gaussians with \( \ell_r^j \), and update their labels to \( \ell_{\text{t}} \).
    
    \item \textit{Whole Match:} This is similar to the Partial Match, but the new part labels come from \( \mathcal{L}_r \). The same comparison process applies.

     \item \textit{New:} If \( \ell_s^i \) cannot be matched with any label in \( \mathcal{L}_r \), we assign a new label to this region.
\end{itemize}

For other cases, we set the labels from \( \ell_s^i \) to the background label (e.g., 0). The \textit{Full Match} is prioritized over other types of matches as it ensures the highest level of consistency.

After label consensus process, we obtain the updated label mapping \( \ell_s \to \ell_t \) for all labels in \( \mathcal{L}_s \), and use it to generate the Consistent Input Label Map. We then update the global Label-Class table \( \mathcal{M}_{LC}^{G} \) based on this mapping and the inpu \( \mathcal{M}_{LC}^{I} \). To enhance consensus accuracy, we also introduce two modules: Input Confidence Update and Part Label Decay.

\textbf{Input Confidence Update. }To avoid incorrect updates of partially visible objects, before label consensus, we update the input confidence \( \mathcal{C}_s \) based on object completeness in the current view. Using the input label map and \( \mathcal{C}_s \), we derive the Input Confidence Map \( \mathcal{I}_s \in \mathbb{R}^{H \times W} \). During GS Voting, the completeness of each \( \ell_r^j \in \mathcal{L}_t \) is calculated as the ratio of visible Gaussians in the current view to the total number in the map. We then calculate the coverage ratio map \( Cov_r \in \mathbb{R}^{H \times W} \) from the rendered label map. Next, we perform element-wise multiplication between \( Cov_r \) and \( \mathcal{I}_s \) to obtain the Integrated Confidence Map. The updated confidence \( \mathcal{C}_s^i \) for each \( \ell_s^i \) is the average of the corresponding values in the Integrated Confidence Map.

\textbf{Part Label Decay. }Several studies \cite{over-sam1,over-sam2,over-sam3} have found that SAM often over-segments objects, which means that a whole object tends to be split into multiple parts. Therefore, we propose a decay mechanism for label confidence. Specifically, whenever a label in a scene is identified as a part label, its confidence is reduced by \( \delta \). This decay ensures that over-segmented labels will be correctly updated over multiple accurate observations.

\begin{figure}
    \vspace{0.6em}
    \centering
    \includegraphics[width=0.48\textwidth]{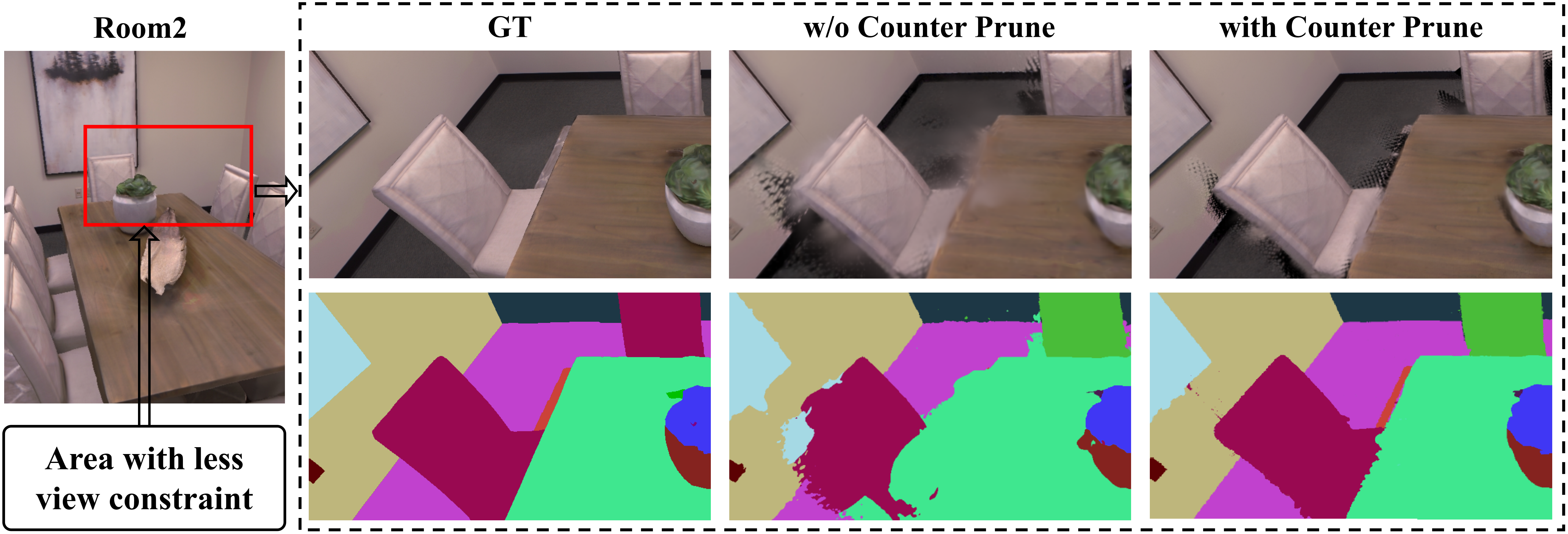}
        \caption{Effect of Segmentation Counter Pruning. \textit{Left:} We select a region with less view constraint throughout the SLAM process. \textit{Right:} Rendered results from new viewpoints with and without the Segmentation Counter Pruning in this region. }
    \label{fig3}
    \vspace{-1.0em}
\end{figure}

\subsection{Segmentation Counter Pruning }
\label{counter}
We observe that in regions with less view constraint, Gaussians can grow excessively large during optimization. This not only prevents the introduction of new Gaussians but also results in inaccurate segmentation. To address this, we propose a Segmentation Counter Pruning method. For fully matched labels \( \ell_s^i \) and \( \ell_r^j \) obtained from the label consensus process, we collect all pixels in the symmetric difference between their respective label maps, denoted as \( S \). Using \(  G_{topK} \), we identify Gaussians in \( S \) with \( \ell_r^j \) as \textit{Counter Gaussians}. If their scale in any direction exceeds a threshold \( \theta \), they are pruned. This method improves segmentation accuracy and allows for the introduction of new Gaussians that better capture the spatial structure, as shown in Fig.\ref{fig3}.

\section{Experiments} 

Given a real-time RGB-D video stream, our method incrementally fuses 2D semantic information from the 2D foundation models to construct a semantic 3D Gaussian scene representation and tracks itself within the scene. Notably, our method does not require any predefined semantic categories. To comprehensively evaluate performance, we focus on both scene understanding and SLAM performance.

For the scene understanding performance, we consider the following factors:

\begin{itemize}
    \item \textit{Multi-view 2D Segmentation}: This metric evaluates the consistency and accuracy of 2D segmentation across multiple viewpoints within the scene. 

    \item \textit{3D Object-Level Understanding}: This evaluates the effectiveness of our semantic 3DGS scene representation in achieving 3D object-level scene understanding.
\end{itemize}

For the SLAM performance, we assess the model’s \textit{camera tracking accuracy} and \textit{reconstruction quality}.

\subsection{Experimental Setup}

\textbf{Datasets:} We evaluate our method on both synthetic and real-world datasets. For scene understanding, we conduct quantitative evaluations on 8 synthetic scenes from Replica\cite{replica}. Following other 3DGS-based SLAM methods, we then assess SLAM performance on the Replica\cite{replica} and TUM\cite{tum} datasets.

\textbf{Baselines:} For scene understanding, we first benchmark against existing Semantic SLAM methods on Replica scenes. We also evaluate our model's capability in open-set scenarios by comparing it with offline methods, including GS-Grouping\cite{Gaussian-Grouping} and Feature 3DGS\cite{Feature-3dgs}. For reconstruction quality and camera tracking, we compare our approach against Visual SLAM methods, including SplaTAM\cite{splatam}, GS-SLAM\cite{gs-slam}, LoopSplat\cite{LoopSplat} and GICP-SLAM\cite{gs-icp}, as well as Semantic SLAM methods like NIDS-SLAM\cite{NIDS}, DNS-SLAM\cite{DNS}, SNI-SLAM\cite{sni}, SGS-SLAM\cite{SGS}, NEDS-SLAM\cite{NEDS} and SemGauss-SLAM\cite{SemGaussSLAM}. 

\textbf{Metrics:} To evaluate scene understanding, we report \textit{mIoU(\%)} and assess pixel-wise segmentation accuracy using \textit{Acc(\%)}. We also measure the frame rate with \textit{FPS} and consider the model's complexity by reporting \textit{Learnable Parameters(MB)}. For reconstruction quality, we use standard photometric rendering metrics such as \textit{PSNR, SSIM}, and \textit{LPIPS}. Camera tracking is evaluated using \textit{ATE RMSE(cm)}. Notably, for \textit{FPS}, we compare the inference time for both the RGB and segmentation maps, as the baselines employ joint rendering.

\textbf{Implementation Details:} Our method is implemented based on GS-ICP SLAM \cite{gs-icp}. We also modified the rasterization component in the official 3DGS code\cite{3dgs} to implement GS Voting Splatting. For the 2D Confidence-based Label Consensus, we set the thresholds as \(\tau_1 = 0.85\), \(\tau_2 = 0.9\), and \(\tau_3 = 0.1\), with the confidence decay parameter \(\delta = 0.06\). The pruning thresholds \( \theta \) for counter Gaussians are set to 0.10 for Replica and 0.25 for TUM. For the 2D foundation models, we tested RAM++\cite{ram}, Yolo-World(v2-l)\cite{Yolo}, MobileSAMv2\cite{Mobilesamv2}, and SAM1(vit-h)\cite{SAM1}, following the default configurations. During experiments, we set the maximum number of labels assignable per test to 2000.

\begin{figure}
     \vspace{0.6em}
    \centering
    \includegraphics[width=0.48\textwidth]{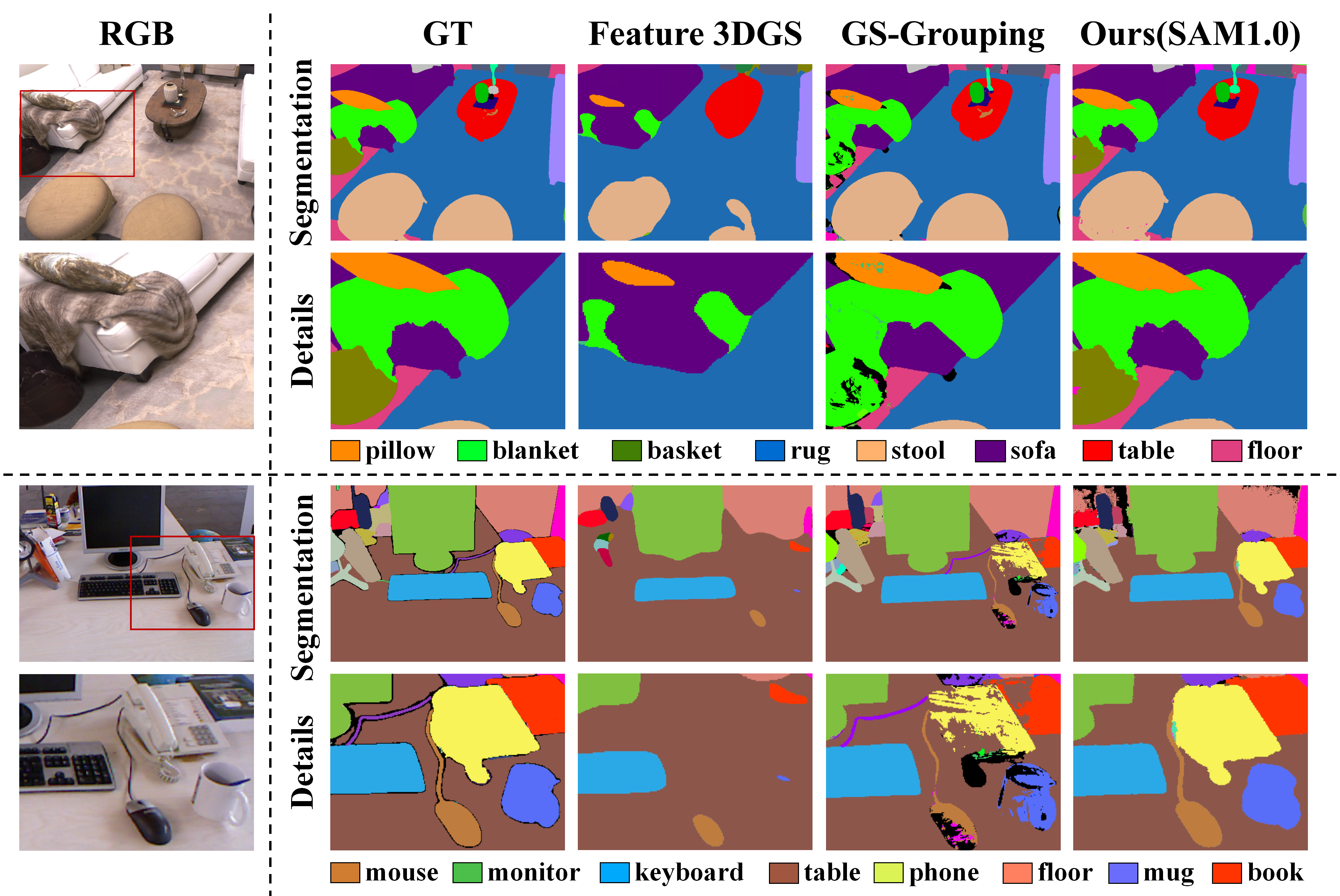}
    \caption{Qualitative comparison of novel-view \textbf{open-set} semantic segmentation. For TUM, novel views refer to viewpoints that are not included in the training data, and the ground truth is obtained from manual annotations.}
    \label{exp1}
    \vspace{-1.0em}
\end{figure}

\subsection{Multi-view 2D Segmentation}
We first compare our method to Radiance-Based semantic SLAM approaches on the Replica dataset, following \cite{SGS, DNS, NIDS, sni,NEDS}, using semantic priors as input to build the map. As shown in \autoref{semantic_accuracy_comparison}, our model achieves the best performance in semantic segmentation accuracy. We attribute this to our superior semantic scene representation, which integrates explicit semantic information into the 3D Gaussian representation, enabling more precise semantic mapping.

However, these methods are restricted to close-set scene understanding.
In contrast, recent offline methods like \cite{Feature-3dgs, Gaussian-Grouping} demonstrated strong open-set scene understanding capabilities by using 2D foundation models and incorporating semantic features into 3DGS representation. We evaluate our method's performance in open-set scenarios by comparing it with these models in zero-shot novel-view semantic segmentation. 
To ensure fairness, we downsampled keyframe data from our SLAM process by a factor of 10 as input and trained the offline models for 6K iterations.

\begin{figure}
    \centering
     \vspace{0.5em}
    \includegraphics[width=0.48\textwidth]{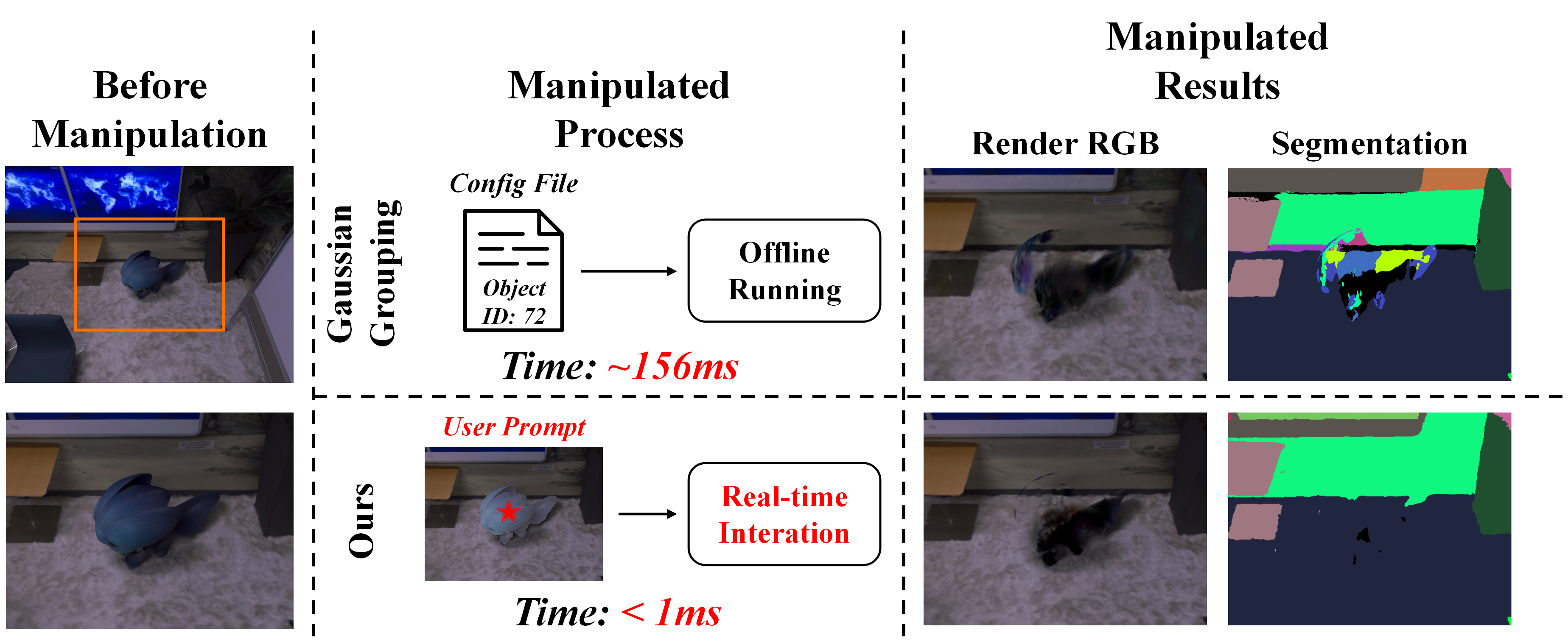}
        \caption{Scene manipulation process and results. \textit{Left:} We select an object from the scene for removal. \textit{Middle:} Compared to GS-Grouping, our model demonstrates more efficient interactive processing. \textit{Right:} Our method achieves superior manipulation results.}
    \label{exp2}
    \vspace{-1.0em}
\end{figure}

\begin{table}[!ht]
\centering
\captionsetup[table]{singlelinecheck=off}
\caption{Quantitative comparison of semantic segmentation accuracy(mIoU↑) against Radiance-Based Semantic SLAM methods on the Replica\cite{replica} datasets. All models use the ground-truth semantic labels from the replica dataset.}
\label{semantic_accuracy_comparison}
\begin{threeparttable}
\setlength{\arrayrulewidth}{1.2pt} 
\resizebox{0.48\textwidth}{!}{
\begin{tabular}{c|cccccccc}
  \toprule
  \textbf{Methods} & R0 & R1 & R2 & Of0 & Of1 & Of2 & Of3 & Of4  \\
  \midrule
  NIDS-SLAM\cite{NIDS} & 82.45 & 84.08 & 76.99 & 85.94 & - & - & - & -  \\
  DNS-SLAM\cite{DNS} & 88.32 & 84.90 & 81.20 & 84.66 & - & - & - & -  \\
  SNI-SLAM\cite{sni} & 88.42 & 87.43 & 86.16 & 87.63 & \cellcolor{lightyellow}78.63 & \cellcolor{lightyellow}86.49 & \cellcolor{lightyellow}74.01 & \cellcolor{lightyellow}80.22 \\
  SGS-SLAM\cite{SGS} & \cellcolor{lightyellow}92.95 & \cellcolor{lightyellow}92.91 & \cellcolor{lightyellow}92.10 & \cellcolor{lightyellow}92.90 & - & - & - & -  \\
  NEDS-SLAM\cite{NEDS} & 90.73 & 91.20 & - & 90.42 & - & - & - & -  \\
  \textbf{Ours(Prior)} & \cellcolor{lightgreen}\textbf{93.24} & \cellcolor{lightgreen}\textbf{94.11} & \cellcolor{lightgreen}\textbf{92.79} & \cellcolor{lightgreen}\textbf{93.22} & \cellcolor{lightgreen}\textbf{92.16} & \cellcolor{lightgreen}\textbf{93.25} & \cellcolor{lightgreen}\textbf{93.14} & \cellcolor{lightgreen}\textbf{93.77}  \\
  \bottomrule
\end{tabular}}
\end{threeparttable}
\vspace{-2em}
\end{table}

\begin{table}[!ht]
\centering
\captionsetup[table]{singlelinecheck=off}
\caption{Comparison of zero-shot novel-view Semantic Segmentation with 3DGS-based \textbf{open-set} scene understanding methods. (average performance on Replica\cite{replica})}
\label{openworld}
\begin{threeparttable}
\setlength{\arrayrulewidth}{1.2pt} 
\resizebox{0.48\textwidth}{!}{
\begin{tabular}{lccccc}
    \toprule
    \textbf{Method} & \textbf{mIoU(\%) ↑} & \textbf{Acc(\%) ↑} &\textbf{\makecell{Render \\ FPS ↑}} & \textbf{\makecell{Learnable \\ Parameters(MB) ↓}}\\
    \midrule
    Featrue 3DGS\cite{Feature-3dgs} & 48.89 & 57.51 & 11.03 & 894.91\\
    GS-Grouping\cite{Gaussian-Grouping} & \cellcolor{lightyellow}59.15 & \cellcolor{lightyellow}69.94 & 16.14 & 717.12\\
    \textbf{Ours(MobileSAMv2)} & 57.48 & 67.14 & \cellcolor{lightyellow}164.78 & \cellcolor{lightyellow}314.11  \\
    \textbf{Ours(SAM1.0)} & \cellcolor{lightgreen}\textbf{61.91} & \cellcolor{lightgreen}\textbf{73.11} & \cellcolor{lightgreen}\textbf{165.47} & \cellcolor{lightgreen}\textbf{301.71} \\
    \bottomrule
\end{tabular}}
\end{threeparttable}
\end{table}

As shown in \autoref{openworld}, our method (using SAM1.0) significantly outperforms the baselines, with \textbf{+13\%} and \textbf{+3\%} mIoU improvements over Feature 3DGS and GS-Grouping, respectively. Similar improvements are visible in Fig.\ref{exp1}. We attribute this to our efficient 2D Confidence-based Label Consensus, which integrates globally consistent semantic information. Moreover, our method uses \textbf{2× fewer} learnable parameters by assigning a 1D \textit{GS Label} to each Gaussian, unlike other methods that use N-channel semantic features. Additionally, our method achieves \textbf{10× faster} rendering speeds, owing to the efficiency of our GS Voting Splatting for rapid semantic label map rendering.

\subsection{3D Object-Level Scene Understanding}
Compared to existing 3DGS-based semantic SLAM methods, our approach not only generates novel-view segmentation results but also supports advanced 3D object-level applications, such as scene manipulation. This includes novel-view synthesis with 3D objects removed, moved, or rotated, as well as interactive scene modifications based on user prompts (e.g., point or bounding box input).

Fig.\ref{exp2} demonstrates that our method achieves more realistic scene manipulation and more efficient user interaction compared to GS-Grouping\cite{Gaussian-Grouping}, due to our advanced semantic 3DGS scene representation.

\begin{figure}
     \vspace{0.5em}
    \centering
    \includegraphics[width=0.48\textwidth]{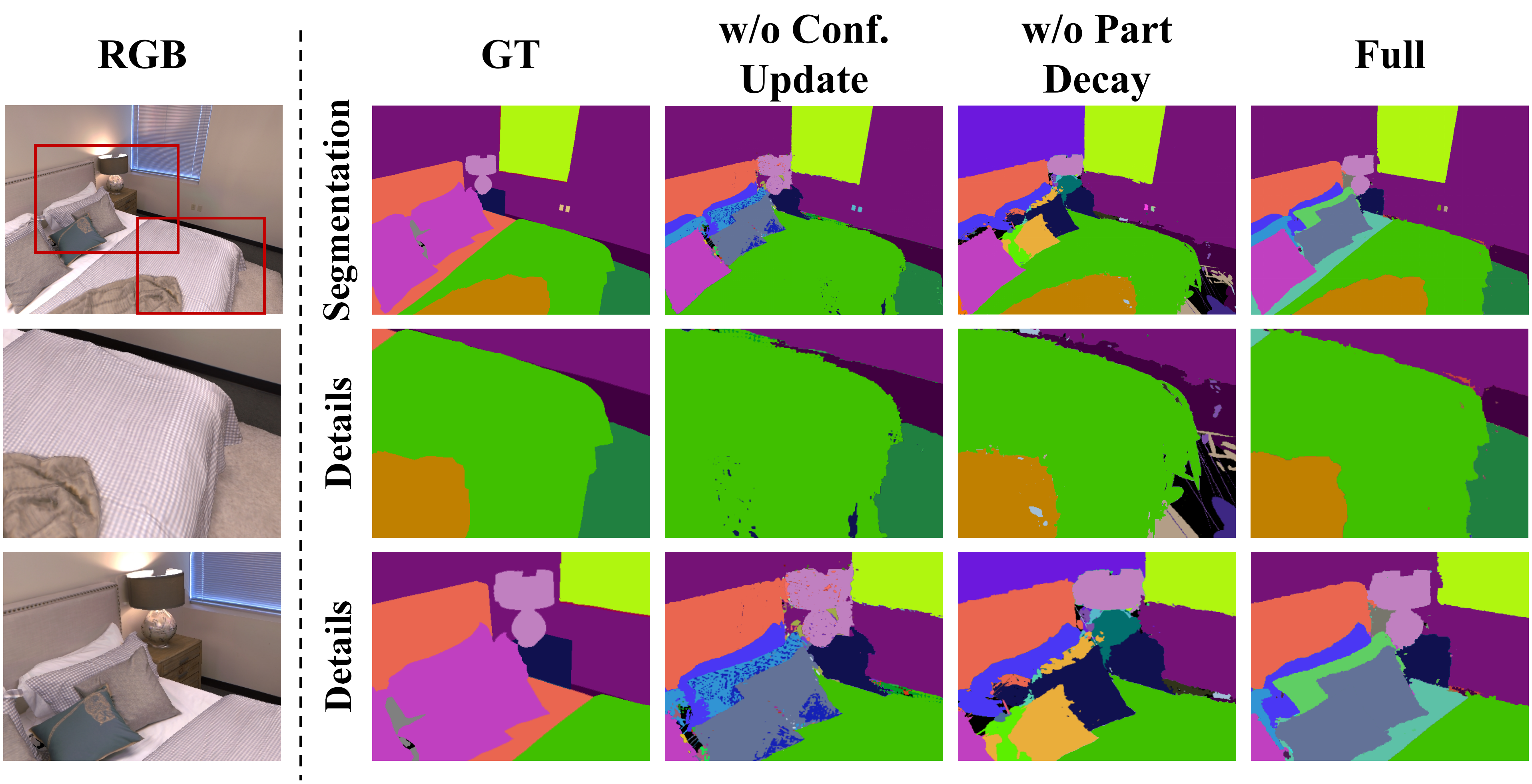}
        \caption{Visualized ablation results. }
    \label{exp3}
    \vspace{-1.0em}
\end{figure}

\subsection{Camera Tracking Accuracy}
As illustrated in \autoref{tab:replica}, our method achieves up to a \textbf{51\%} relative improvement in tracking accuracy compared to other dense semantic SLAM methods on the Replica dataset. This improvement is attributed to the integration of the G-ICP tracking module derived from GICP-SLAM. Additionally, we also achieve the best localization performance on the TUM dataset, as shown in \autoref{tab2}. We attribute this improvement to the Segmentation Counter Pruning strategy, which enhances scene representation by refining object contours. As a result, the alignment of Gaussian distributions during the G-ICP process is significantly improved, leading to superior localization accuracy.

\subsection{Reconstruction Quality} As shown in \autoref{tab:replica} in the Replica scene, our method achieves the best performance in PSNR and LPIPS compared to other dense semantic SLAM and visual SLAM methods.

\begin{table}[!ht]
\centering
\captionsetup[table]{singlelinecheck=off}
\caption{Camera Tracking and Reconstruction Results on Replica\cite{replica}. (average performance on 8 scenes)}
\label{tab:replica}
\begin{threeparttable}
\large
\setlength{\arrayrulewidth}{1.2pt} 
\resizebox{0.45\textwidth}{!}{
\begin{tabular}{c|l|cccc}
  \toprule
  \textbf{Category} & \textbf{Methods} & \textbf{ATE↓} & \textbf{PSNR↑} & \textbf{SSIM↑} & \textbf{LPIPS↓} \\
  \midrule
  \multirow{3}{*}{\makecell{Visual \\ SLAM}} 
  & SplaTAM\cite{splatam}  & 0.35 & 33.91 & 0.969 & 0.097 \\
  & GS-SLAM\cite{gs-slam} & 0.50 & 34.27 & 0.975 & \cellcolor{lightyellow}0.082 \\
  & LoopSplat\cite{LoopSplat} & \cellcolor{lightyellow}0.26& \cellcolor{lightyellow}36.63 & \cellcolor{lightgreen}\textbf{0.985} & 0.112 \\
  & GICP-SLAM\cite{gs-icp} & \cellcolor{lightgreen}\textbf{0.16} & \cellcolor{lightgreen}\textbf{38.86} & \cellcolor{lightyellow}0.976 & \cellcolor{lightgreen}\textbf{0.041} \\
  \midrule
  \multirow{5}{*}{\makecell{Semantic \\ SLAM}} 
  & SNI-SLAM\cite{sni} & 0.46 & 29.43 & 0.935 & 0.235 \\
  & SGS-SLAM\cite{SGS} & 0.41 & 34.66 & 0.973 & 0.096 \\
  & NEDS-SLAM\cite{NEDS} & 0.35 & 34.76 & 0.962 & 0.088 \\
  & SemGauss-SLAM\cite{SemGaussSLAM} & \cellcolor{lightyellow}0.33 & \cellcolor{lightyellow}35.03 & \cellcolor{lightgreen}\textbf{0.982} & \cellcolor{lightyellow}0.062 \\
  & \textbf{Ours (SAM1.0)} & \cellcolor{lightgreen}\textbf{0.16} & \cellcolor{lightgreen}\textbf{39.49} & \cellcolor{lightyellow}0.978 & \cellcolor{lightgreen}\textbf{0.034} \\
  \bottomrule
\end{tabular}}
\end{threeparttable}
\end{table}

\begin{table}[!ht]
 \vspace{0.5em}
\centering
\captionsetup[table]{singlelinecheck=off}
\caption{Camera Tracking Results on Tum\cite{tum}. ATE.↓}
\label{tab2}
\begin{threeparttable}
\large
\setlength{\arrayrulewidth}{1.2pt} 
\resizebox{0.45\textwidth}{!}{
\begin{tabular}{c|l|cccc}
  \toprule
  \textbf{Category} & \textbf{Methods} & \textbf{fr1-desk} & \textbf{fr2-xyz} & \textbf{fr3-office} & \textbf{Avg.} \\
  \midrule
  \multirow{3}{*}{\makecell{Visual \\ SLAM}} 
  & SplaTAM\cite{splatam}
      & 3.35 
      & 1.24 
      & 5.16 
      & 3.25 \\ 
  & GS-SLAM\cite{gs-slam}
      & 3.65 
      & - & - & - \\
  & LoopSplat\cite{LoopSplat}
      & \cellcolor{lightgreen}\textbf{2.08} 
      & \cellcolor{lightyellow}1.58 
      & 3.22
      & 2.29 \\
  & GICP-SLAM\cite{gs-icp}
      & 2.41 
      & 1.77 
      & \cellcolor{lightyellow}2.67 
      & \cellcolor{lightyellow}2.28 \\
  \midrule
  \multirow{1}{*}{\makecell{Semantic SLAM}} 
  & Ours (SAM1.0)  
      & \cellcolor{lightyellow}2.40 
      & \cellcolor{lightgreen}\textbf{1.57} 
      & \cellcolor{lightgreen}\textbf{2.48} 
      & \cellcolor{lightgreen}\textbf{2.15} \\ 
  \bottomrule
\end{tabular}}
\end{threeparttable}
 \vspace{-1.0em}
\end{table}

\subsection{Ablation Study. }
To better validate our method, we conduct ablation studies on the Replica datasets. We primarily investigate the impacts of our proposed Segmentation Counter Pruning in \ref{sec:concensus} and the Input Confidence Update, Part Label Decay in \ref{counter}. Using our(SAM1.0) as the baseline, we compare segmentation performance after removing each module. The quantitative and qualitative results are presented in \autoref{ablation} and Fig.\ref{exp3}, respectively. \textbf{1) }Part Label Decay: This module boosted the segmentation results by 5.96↑ mIoU\% and 10.2↑ Acc\%, as it ensures over-segmented regions are refined after multiple observations, leading to more compact segmentation. \textbf{2) }Confidence Update: This module improved results with 3.39↑ mIoU\% and 5.4↑ Acc\%, as it prevents incorrect updates for partially observed objects during SLAM. \textbf{3) } Segmentation Counter Pruning: This module yielded a smaller improvement of 0.26↑ mIoU\% and 1.49↑ Acc\%, since it primarily refines object boundaries rather than enhancing overall segmentation performance.

\begin{table}[htbp]
    \centering
    \captionsetup[table]{singlelinecheck=off}
    \caption{Segmentation Performance comparison of various settings on Replica\cite{replica}.}
    \label{ablation}
    \setlength{\arrayrulewidth}{1.2pt} 
    \resizebox{0.46\textwidth}{!}{%
    \begin{tabular}{lccc}
        \toprule
        \textbf{Settings} & \textbf{mIoU(\%) ↑} & \textbf{Acc(\%) ↑}   \\
        \midrule
        w/o Segmentation Counter Pruning & 62.91 & 73.02 \\
        w/o Input Confidence Update & 59.78 & 69.11  \\
        w/o Part Label Decay & 57.21 & 64.31  \\
        \midrule
        Ours(SAM1.0) & \textbf{63.17} & \textbf{74.51}  \\
        \bottomrule
    \end{tabular}%
    }
\end{table}

\section{CONCLUSIONS}

We propose OpenGS-SLAM, a novel framework that uses 3D Gaussian representation for dense semantic SLAM in open-set scenarios. By integrating semantic labels into the 3D Gaussian model, it enables efficient 3D object-level understanding and dense semantic mapping. We introduce Gaussian Voting Splatting for rapid label map rendering and scene updates. To ensure consistency in multi-view semantic labels from SAM, we propose Confidence-based 2D Label Consensus. Additionally, we implement Segmentation Counter Pruning to refine the scene representation. Extensive experiments show that our approach achieves state-of-the-art performance in tracking, mapping, and scene understanding. However, our method is not applicable to dynamic scenes. In the future, we aim to leverage semantic information to achieve object-level understanding in dynamic environments. We also plan to develop a well-annotated large-scale RGB-D dataset, which includes common and uncommon objects, to further validate the model's performance in open-set environments.










\bibliographystyle{ieeetr}
\bibliography{reference}

\end{document}